# Hey Robot, Which Way Are You Going? Nonverbal Motion Legibility Cues for Human-Robot Spatial Interaction*

Nicholas. J. Hetherington, Elizabeth. A. Croft and H. F. Machiel Van der Loos

*Abstract*—Mobile robots have recently been deployed in public spaces such as shopping malls, airports, and urban sidewalks. Most of these robots are designed with human-aware motion planning capabilities but are not designed to communicate with pedestrians. Pedestrians that encounter these robots without prior understanding of the robots' behaviour can experience discomfort, confusion, and delayed social acceptance. In this work we designed and evaluated nonverbal robot motion legibility cues, which communicate a mobile robot's motion intention to pedestrians. We compared a motion legibility cue using Projected Arrows to one using Flashing Lights. We designed the cues to communicate path information, goal information, or both, and explored different Robot Movement Scenarios. We conducted an online user study with 229 participants using videos of the motion legibility cues. Our results show that the absence of cues was not socially acceptable, and that Projected Arrows were the more socially acceptable cue in most experimental conditions. We conclude that the presence and choice of motion legibility cues can positively influence robots' acceptance and successful deployment in public spaces.

## I. INTRODUCTION

Mobile robots are well-accepted in controlled environments either with trained operators or separate from humans. These mobile robots are increasingly being deployed in public spaces such as airports, shopping malls, and urban sidewalks. These robots provide economic and social benefits to their owners and end users but can be a disruption to others in society when not designed well. Our research focuses on the design of mobile robots for human-robot spatial interaction (HRSI) in public spaces.

For most pedestrians, human-human interaction in public spaces is natural and fluid. However, this interaction relies on an understanding of how others move [1]. Pedestrians inherently assume that other pedestrians will cooperate in mutual collision avoidance. Furthermore, natural crowd movement relies on pedestrians' subtle body language cues [2]. Mobile robots are new, unfamiliar agents that have not yet been incorporated into public interactions. Delivery and service robots are designed for the tasks at the endpoints of their journeys through a public space, such as interacting with restaurant owners and customers, or transporting airport luggage. These robots are designed to move safely and efficiently through the environment, but, apart from simple visual and/or audio alarms, they do not communicate with pedestrians along the way. As a result, more and more pedestrians are encountering spatial interactions with these robots without prior understanding of their behaviour.

Most research in HRSI contexts focuses on human-aware navigation and motion planning. Many researchers design robot motion that adheres to social conventions in public spaces. However, most approaches to human-aware navigation do not include robot legibility. In addition to human awareness, legible robot behaviour is a key component of HRSI [3].

In this paper we design and evaluate nonverbal robot motion legibility cues, which communicate a mobile robot's motion intentions in human-robot spatial interactions. Our specific contribution identifies which of two genres of cue is more socially acceptable. We also explore which type of information each cue should communicate and consider multiple Robot Movement Scenarios. Fig. 1 shows the robot motion legibility cues. By increasing robots' motion legibility, we can design more socially acceptable robots for public spaces.

## II. RELATED WORK

Past research has investigated a variety of communication modalities for robot-to-human cues. St. Clair and Mataric used verbal feedback to coordinate a human and mobile robot in a collaborative task [4]. Thomas *et al.* used beeping and coloured LEDs to indicate a state change in their mobile robot task planner for HRSI at doorways [5]. These studies show sound is a good modality for infrequently communicating high-level task information. Motion legibility cues, however, require frequent communication of more

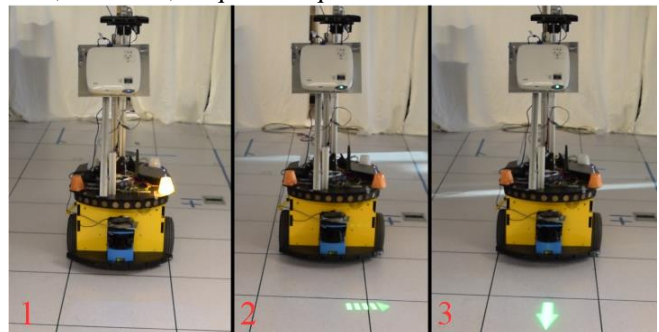

Fig. 1. The robot motion legibility cues: (1) Flashing Lights; (2) Projected Arrows in Goal mode; (3) Projected Arrows in Path mode.

* Research supported by the Natural Sciences and Engineering Research Council of Canada, the Government of British Columbia, and Australian Research Council Discovery Project ID DP200102858.

N. J. Hetherington and H.F.M. Van der Loos are with the Department of Mechanical Engineering, University of British Columbia, Canada (nickjh@alum.ubc.ca, vdl@mech.ubc.ca).

E. A. Croft is with the Departments of Mechanical and Aerospace Engineering and Electrical and Computer Systems Engineering at Monash University, Australia (elizabeth.croft@monash.edu).

detailed motion information and are less understandable when using sound.

Fischer *et al.* designed two gazing methods for a service robot with an anthropomorphic face [6]. Their results showed the robot should use gaze to communicate attention to pedestrians, not as a motion legibility cue. May *et al.* compared robot gaze to flashing lights as motion legibility cues for mobile robots [7]. They found the flashing lights to be more communicative and make the participants more comfortable. The literature does not show robot gaze to be an effective motion legibility cue for mobile robots.

Shrestha *et al.* compared arrows on a display screen to flashing lights as motion legibility cues in a head-on interaction between a walking human and mobile robot [8]. The flashing lights indicated the robot would move to the left or right to avoid the human, and the arrows indicated the human should move to one side or the other. In two other conditions these cues were combined with the sound of a motor vehicle turning signal. Subjects rated the flashing lights cue higher than the display screen cue in terms of comfort, naturalness, performance, and predictability. The addition of sound had no positive effect on either cue.

The results from [7] and [8] show that flashing lights are a more communicative motion legibility cue than gaze or display screens. The robots in [7] and [8] use two fixed flashing lights, which are limited to on/off states and use a single colour. Szafir *et al.* used an LED strip to communicate a quadrotor robot's intended direction [9]. LED strips, featuring tens of LEDs with different colours, have the potential to communicate more complex information than pairs of flashing lights. However, pairs of flashing lights benefit from the mental models of existing cues in society: turning signals on motor vehicles.

Multiple works have investigated robot motion legibility cues that use light projection systems. Shrestha *et al.* projected a red arrow onto the ground to communicate a mobile robot's intended direction of travel [10]. Results showed the projection cue increased the robot's legibility and subjects' comfort compared to no communication. As in their previous study, the addition of sound had no positive effect on the projection cue. Matsumaru *et al.* projected a line on the ground to communicate a mobile robot's direction and speed [11]. Participants rated this communication cue as generally understandable but commented that arrows might be better than a simple line. Coovert *et al.* projected arrows and a simplified map onto the ground to show a mobile robot's intended movement [12]. Their field study showed that projected arrows can communicate both short and long-term intended movement. Chadalavada *et al.* designed a projection-based motion legibility cue for a robotic forklift [13]. Subjects rated the projection cue significantly higher than the no-cue control condition in terms of communication, reliability, predictability, transparency, and situational awareness. Watanabe *et al.* designed a projection-based motion legibility cue for a robotic power wheelchair and evaluated it in a head-on interaction with a walking pedestrian and a passenger sitting in the moving wheelchair [14]. Both the pedestrians and the passengers found the projection cue increased their comfort and the robot's motion legibility.

In summary, the reviewed literature shows that flashing lights and projection systems are communicative motion legibility cues. The other methods reviewed above – sound, robot gaze, and display screens – are less effective as cues. Despite the interest in these motion legibility cues, we were not able to find a direct comparison of flashing lights to projection systems. Flashing lights and projection systems are also fundamentally different. Projection systems are a form of spatial augmented reality extended beyond the robot [12], whereas flashing lights are fixed to the robot. Flashing lights use on/off states to communicate, whereas projections can use different shapes. Both can use colours and frequencies to communicate. Furthermore, flashing lights are prevalent in society, whereas projection systems are less familiar. These differences and the lack of direct contrast in the research warrant a comparison of these two promising cue modalities, which we call Cue Types below.

Multiple works in HRSI differentiate between two motion legibility factors: path-predictability, in which the human understands the robot's next immediate movement, and goal-predictability, in which the human understands the robot's next intermediate destination in the world. Using different terminology and in the context of robot manipulators, Dragan *et al.* distinguish between path- and goal-predictability as "fundamentally different and often contradictory properties of motion" [15]. Lichtenthäler and Kirsch analyzed a path-crossing human-robot spatial interaction to test Dragan *et al.*'s claim in the context of mobile robotics [16]. They found a correlation between path- and goal-predictability, which differs from Dragan *et al.*'s findings. However, [15] and [16] used only the robot's motion to communicate, no additional visual cues. These contradictory findings and the lack of research with visual cues warrant exploring whether visual motion legibility cues should be designed for path- or goal-predictability, which we call Cue Modes below.

III. DESIGN AND FUNCTION OF MOTION LEGIBILITY CUES

We designed the robot motion legibility cues to explore the two research gaps identified in Section II. To fill the research gap in comparing the two most successful communication modalities from the literature, we designed two Cue Types: Projected Arrows and Flashing Lights. The Projected Arrows display a green arrow on the ground in front of the robot. The Flashing Lights use an orange LED on either side of the robot. To explore the research gap relating to the type of information these cues should communicate, we designed three Cue Modes: Path mode, Goal mode, and Path&Goal mode. Path&Goal mode is a sequential combination of Path mode and Goal mode. Each Cue Type can operate in any of the three Cue Modes.

We designed the motion legibility cues using a series of pilot studies to optimize the designs before comparing them to each other. We identified promising designs from the literature for each cue and tested them with 10 participants. We determined that the Flashing Lights should use a pair of separated LEDs rather than a continuous LED strip. We valued the simplicity of a separated pair of LEDs and wanted to leverage existing mental models from motor vehicle turning signals. We piloted different colours for each Cue Type. Pilot data supported orange Flashing Lights to match motor vehicle turning signals. This choice is supported by [7] and [8]. Pilots supported green Projected Arrows because it is the "friendliest colour" and "suggests motion". Pilots indicated that simple arrows were more understandable than a curving, full-length path. Furthermore, the limited size of the projection image only allowed a scaled-down path, which confused participants. This choice is supported by [10]–[12]. Path mode communicates immediately forthcoming motion; Goal mode communicates major waypoints in a structured environment. In Goal mode, pilots showed that the robot should communicate short-range waypoints in the human's field of view, rather than long-range waypoints out of sight.

Pilot studies showed that variable frequency can be used to communicate temporal information. Sharper angles in the robot's path occur faster, so we varied the Flashing Lights' frequency in Path mode to communicate robot's forthcoming heading. The Projected Arrows inherently communicate this angle, so they did not flash in Path mode. In Goal mode both Cue Types use frequency to represent predicted time to the intermediate goal. Pilot studies also showed that the fill of the Projected Arrows could also communicate temporal information – we used a solid, constant arrow to communicate immediate motion in Path mode and a dashed, flashing arrow to communicate future motion in Goal mode. Fig. 2 and Table 1 show how the cues operate.

We developed the robot motion legibility cues on a PowerBot differential drive mobile base (Adept Mobile Robots) using the Robot Operating System (ROS). The robot is 83 cm long, 63 cm wide, and 173 cm tall. We used the open-source ROS Navigation stack to move the robot along planned paths between scripted intermediate goals in the environment. The ROS package we created for this research is documented in [17] and available on GitHub[1].

## IV. USER STUDY

We designed a user study to test two hypotheses: (H1) Projected Arrows are a more socially acceptable Cue Type than Flashing Lights, and (H2) Path&Goal mode is a more socially acceptable Cue Mode than either Path mode or Goal mode. We also tested whether participants comprehended the intended information from the different Cue Modes and explored two Robot Movement Scenarios. The user study was approved by the University of British Columbia Behavioural Research Ethics Board.

### A. Experimental Scenario

To facilitate data collection, we captured videos of the robot's motion legibility cues as it moved through the structured environment shown in Fig. 2. The passageway and junction (Goal 1) created visual structural context. In each video the robot first moved around the obstacle to the junction, then either straight ahead or to the robot's left (Goal 2). This created two different Robot Movement Scenarios. Avoiding the obstacle created more diverse motions to further demonstrate the cues. We recorded videos from the point of view of the stationary observer shown in Fig. 2. As suggested by pilot study participants, we paused the robot for 5 seconds at the junction to allow the observer to focus on and process the motion legibility cue. During the pause the next goal, or the path thereto, was used to animate the robot's motion legibility cues as a motion preview.

To achieve repeatable robot movement in each video, we initially teleoperated the robot through each Scenario (Turn, Straight). While running the ROS Navigation stack we recorded the following data: velocity control (teleoperated), path (planned by ROS Navigation), and goal poses (sent to ROS Navigation). We then input these data to the robot controller and the cue controller while we recorded each video. Copies of the videos are archived in [17]; a compilation is attached to this paper and available on YouTube[2].

### B. Survey Design and Experimental Procedure

We designed a survey to collect data online using the videos described above. After consenting and answering demographic questions, participants saw instructions explaining the situational context of the videos. The instructions included an image of the experimental scenario. Participants then viewed a control video with no motion legibility cue in both Robot Movement Scenarios (Turn or Straight). Participants were then grouped by Scenario. Each group was shown 6 videos in randomized order – one for each Cue Type (Projected Arrows, Flashing Lights) operating in each Cue Mode (Path, Goal, Path&Goal). As an attention-check question, participants confirmed the robot's finish location. After each video, including the control videos, participants responded to these statements on a 5-point Likert scale:

1.1. The robot's communication to me was clear.
1.2. The robot moved as I expected.
1.3. The robot's communication showed me its next movement.
1.4. As the robot approached the junction in the middle of the scene, its final destination was clear.
1.5. The robot's overall behaviour was reasonable.

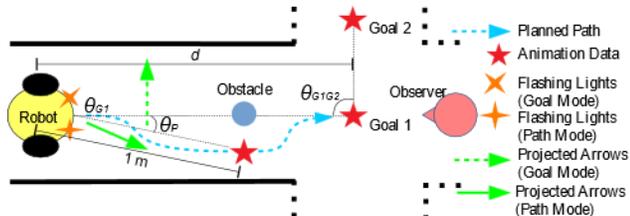

Fig. 2. Cue function and experimental setup for the user study. One pair of Cue Type and Cue Mode activated in each trial. Black lines represent passage borders. Arrows were 30 cm. Table 1 shows the variables' usage.

Table 1. Animation data, flashing frequencies, and activation conditions for each Cue Type and Cue Mode. Variables are shown in Fig. 2 and are all in the robot's coordinate frame: $\theta_p$ is the angle to the point 1 m along the planned path; $\theta_{G1}$ is the angle to Goal 1. $\theta_{G1G2}$ is the angle between Goal 1 and Goal 2. $d$ is the distance to Goal 1. We limited all frequencies to 0.5-5 Hz to ensure visibility.

**Flashing Lights**

| Mode | Data | Flashing Freq. | Activation Condition |
|---|---|---|---|
| Path | $\theta_p$ | 0.1 $\theta_p$ | $\theta_p > 20°$ |
| Goal | $\theta_{G1G2}$ | $5d$ | $d < 1.5$ m AND $\theta_{G1} < 45°$ AND $\theta_{G1G2} > 20°$ |

**Projected Arrows**

| Mode | Data | Flashing Freq. | Activation Condition |
|---|---|---|---|
| Path | $\theta_p$ | None | Always On |
| Goal | $\theta_{G1G2}$ | $5d$ | $d < 1.5$ m AND $\theta_{G1} < 45°$ |

[1] https://github.com/njhetherington/commbot

[2] https://youtu.be/BQUwUYU2pPA

1.6. The robot's communication would be socially compatible in a pedestrian's environment.
1.7. The robot's communication made me feel comfortable.
1.8. I liked the robot.

We used a Social Acceptability Scale comprising the Likert-type Statements 1.1, 1.2, and 1.5-1.8. We calculated the Social Acceptability Score as the mean response to the statements above. Joshi *et al.* support taking the mean of a set of Likert-type items to create interval scale data [18]. We analyzed the internal reliability of the Social Acceptability Scale with Cronbach's α for each combination of Cue Type, Cue Mode, and Robot Movement Scenario. The results showed a minimum α of 0.86, above the minimum of 0.70.

We separately designed Statements 1.3 and 1.4 to directly assess comprehension of Path mode and Goal mode. In Path mode or Path&Goal mode, higher responses to 1.3 than to 1.4 show comprehension of the robot's path. In Goal mode or Path&Goal mode, higher responses to 1.4 than to 1.3 show comprehension of the robot's goal.

At the end of the survey, participants answered three follow-up questions, FQ1-3. FQ1 used a slider to measure preference for Projected Arrows or Flashing Lights. FQ2 used three statements and a 5-point Likert scale to assess the identifiability of the three Cue Modes when demonstrated with either Projected Arrows or Flashing Lights. FQ3 used a slider to measure preference among the three Cue Modes.

We piloted our online survey with 8 participants before the full experiment. After minor adjustments, we found participants understood the instructions, videos, and survey questions. The average length of the full study was 36 mins.

We recruited participants using Amazon Mechanical Turk (MTurk) for a reward of US $2.50 and on social media for no reward. A total of 289 participants responded to the online survey – 255 on MTurk and 34 elsewhere. We excluded 60 participants due to one or more incorrect responses to the attention-check question. Of the 229 remaining participants, 81 identified as female, 146 as male, 1 as non-binary, and 1 preferred not to say. They placed themselves in age brackets: 16% were ages 18-25, 21% were 26-30, 33% were 31-40, 16% were 41-50, 9% were 51-60, and 5% were over 60. Participants also described their prior robotics experience, for which none were excluded.

*C. Data Analysis*

The No Cue control condition in the first two videos does not apply to the Cue Mode factor. We therefore analyzed the Social Acceptability Scale (SAS) data with two separate mixed-model analysis of variance (ANOVA) tests. We performed a 3x2 mixed ANOVA on the SAS data including the No Cue condition; the within-subjects factor was Cue Type (Projected Arrows, Flashing Lights, No Cue) and the between-subjects factor was Robot Movement Scenario (Turn, Straight). We also performed a 2x3x2 mixed ANOVA on the SAS data *excluding* No Cue condition; the within-subjects factors were Cue Type (Projected Arrows, Flashing Lights) and Cue Mode (Path, Goal, Path&Goal), and the between-subjects factor was Robot Movement Scenario. Due to the large sample size and the lack of extreme outliers, the mixed ANOVAs were robust to violations of the tests' assumptions. To assess *Cue Mode comprehension*, we analyzed data from Statements 1.3 and 1.4 and from FQ2 using three separate nonparametric Aligned Rank Transform (ART) ANOVAs [19]. For all four ART ANOVAs, the within-subjects factors were Cue Type (Projected Arrows, Flashing Lights) and Cue Mode (Path, Goal, Path&Goal), and the between-subjects factor was Robot Movement Scenario (Turn, Straight). We followed each ANOVA with planned pairwise comparisons using the Bonferroni correction. Because we performed two ANOVAs on the SAS data, we did not analyze effects with significance levels below a Bonferroni-corrected α of 0.25. FQ1 and FQ3 did not allow for statistical analysis, so we performed a graphical analysis using boxplots.

V. RESULTS

The mixed ANOVA on Social Acceptability Score *excluding* the No Cue control data detected a significant three-way interaction between Cue Type, Cue Mode, and Robot Movement Scenario ($F(2,434) = 32.1$, $p < .001$, $\eta_p^2 = 0.12$). Fig. 3 illustrates the results. The results *do not* show statistical support for H1 in every pairwise comparison, but they do support the selection of Projected Arrows over Flashing Lights. In the Straight Scenario, Projected Arrows scored statistically higher than Flashing Lights when operating in all three Cue Modes. In the Turn Scenario, Projected Arrows scored statistically higher only when operating in Path Mode. The graphical analysis of FQ1 showed a preference for Projected Arrows over Flashing Lights. Similar to the Social Acceptability Score data, this preference was stronger in the Straight Scenario than in the Turn Scenario. The results *do not* show statistical support for H2, but they do reveal significant differences in the Turn Scenario. When using Projected Arrows, the statistically significant order of social acceptability is: (1) Path mode; (2) Path&Goal mode; (3) Goal mode. When using Flashing Lights, the statistically significant order is: (1) Path&Goal mode; (2) Goal mode; (3) Path mode. The graphical analysis of FQ3 showed a preference for Path&Goal mode in both Projected Arrows and Flashing Lights.

The mixed ANOVA on the SAS data *including* the No

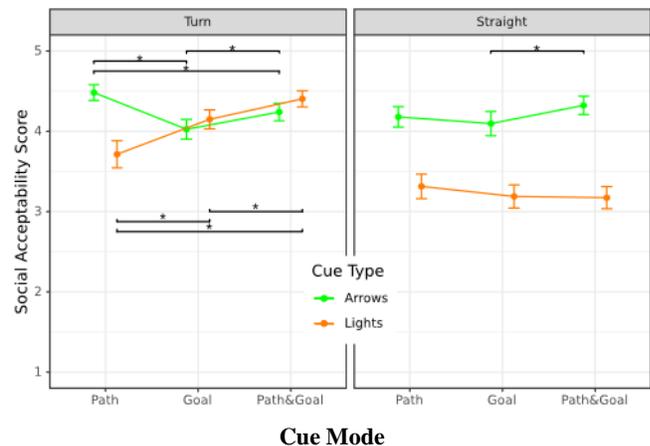

**Social Acceptability of Robot Motion Legibility Cues**

Fig. 3. Social Acceptability Scores for each Cue Type when operating in each Cue Mode, divided by Robot Movement Scenario. The coloured confidence intervals around each mean are within-subjects 95%. The black brackets show statistically significant differences with $p < .05$. ($N$=229)

Cue control condition detected a significant two-way interaction between Cue Type and Robot Movement Scenario ($F(2,407) = 26.5$, $p < .001$, $\eta_p^2 = 0.11$). Planned pairwise comparisons revealed that in the Turn Scenario the No Cue condition was significantly less socially acceptable than both the Projected Arrows ($M = 0.9$, $p < .001$, $r = 0.45$) and the Flashing Lights ($M = 0.7$, $p < .001$, $r = 0.38$). In the Straight Scenario, the Flashing Arrows were significantly more socially acceptable than the No Cue condition ($M = 1$, $p < .001$, $r = 0.47$), but the Flashing Lights scored only slightly higher than the No Cue condition.

The ART ANOVAs on *comprehension* of Cue Modes detected significant three-way interactions between Cue Type, Cue Mode, and Robot Movement Scenario in both the Statement 1.3 data ($F(2,454) = 29.3$, $p < .001$, $\eta_p^2 = 0.11$) and the Statement 1.4 data ($F(2,454) = 27.8$, $p < .001$, $\eta_p^2 = 0.11$). Fig. 4 illustrates the results. As noted in Section IV.B: in *Path* mode or *Path&Goal* mode, higher responses to 1.3 than to 1.4 show comprehension of the robot's *path*. In Goal mode or Path&*Goal* mode, higher responses to 1.4 than to 1.3 show comprehension of the robot's *goal*. By this assessment, correct Cue Mode comprehension is demonstrated in 20 of 24 comparisons. The other four comparisons show a lack of comprehension; for example, in the Turn scenario with the Lights in Path mode, responses to Statement 1.3 were *lower* in Path mode than in Goal mode (top left quadrant of Fig. 4). Fig. 4 also shows that Cue Mode comprehension in the Straight Scenario is much higher when using Projected Arrows than when using Flashing Lights.

The ART ANOVA on the *identifiability* of Cue Modes (FQ2) detected a significant three-way interaction between Cue Type, Cue Mode, and Robot Movement Scenario ($F(2,454) = 35.2$, $p < .001$, $\eta_p^2 = 0.11$). Pairwise comparisons revealed the order of Cue Mode identifiability was: (1) Path mode; (2) Goal mode; (3) Path&Goal mode, although not all the differences were statistically significant. Further details on methods and results can be found in [17].

## VI. DISCUSSION

Our direct comparison between Flashing Lights and Projected Arrows is novel, as is the comparison of Cue Modes with visual motion legibility cues. The most socially acceptable cue was Projected Arrows in Path mode. Overall, participants preferred Projected Arrows over Flashing Lights. However, the results lead to more nuanced design recommendations. In the Turn Scenario there are statistically significant differences that inform the choice of Cue Mode for each Cue Type: Flashing Lights score higher in Goal mode and Path&Goal mode than they do in Path mode; Projected Arrows score higher in Path mode and Path&Goal mode than they do in Goal mode. Path&Goal mode communicates both path and goal information, so it follows that Projected Arrows best communicate path information, and Flashing Lights best communicate goal information. We also see Flashing Lights are best suited to the Turn Scenario, whereas Projected Arrows are effective in both Turn and Straight Scenarios. The cues animate based on basic heading and goal data from the robot's navigation system, so they can be generically applied to autonomous mobile robots.

Our finding that Flashing Lights should operate in Goal mode supports May *et al.*'s results, where lights communicated goal information [7]. However, this finding contradicts [8], where lights communicated path information. Our work extends [7] and [8] to show what type of information these cues should communicate. The success of our Path&Goal mode shows support for Lichtenthäler and Kirsch's finding that path- and goal-predictability are not contradictory in a mobile robotics context [16]. Coovert *et al.* showed projected arrows can communicate short- and long-term motion in separate trials [12]. The success of our Path&Goal mode shows that Projected Arrows can communicate different information sequentially.

Flashing Lights in Goal mode remain off in the Straight Scenario to match motor vehicle turn signals. Pilots also showed that deactivating the lights when moving straight helped draw attention to the lights when the robot signaled a turn. Despite this difference, the Cue Types are comparable. The robot's forward motion is still communicative when the Flashing Lights are off, and both Cue Types animate during turns. We designed the cues based on both the literature and multiple pilot tests and optimized the cues before rigorously comparing them. Flashing Lights are well-accepted, so it is valuable to test them against a less-studied Projected Arrows.

However, the cues' complex design may limit the generalizability of our results. The experimental videos show some unexpected artifacts that may have created confounds. In some videos, the robot started turning before the cue updated. In others, the dynamic nature of the robot's planned path caused the lights to flash unexpectedly. While these artifacts may limit internal validity, they may also be representative of a less-controlled real world.

The Projected Arrows face practical limitations in the real world: occlusions by the environment and the robot itself; and lowered visibility in sunlight.

Experimentation with videos and an online survey is a limitation of the user study. The COVID-19 pandemic interrupted plans for in-person trials. MTurk data are less trustworthy than in-person data. We used post-video attention-check questions to exclude low-quality responses,

**Comprehension of Robot Motion Legibility Cues**

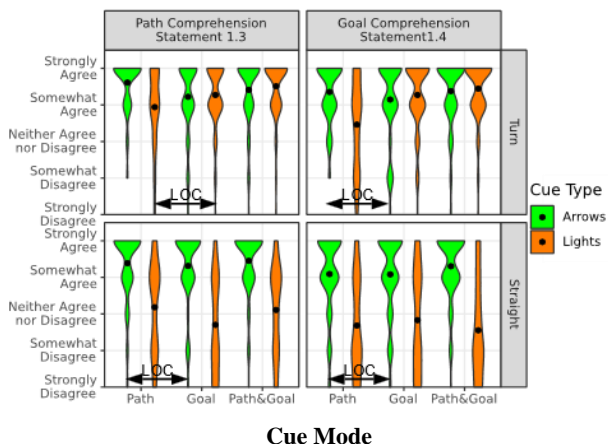

Fig. 4. Data from Statements 1.3 and 1.4 with different Cue Modes, divided by Cue Type and Scenario. Black points show means on a 5-point Likert scale to help distinguish the violins. Brackets show a lack of comprehension (LOC) of Cue Modes; all other comparisons show comprehension. ($N$=229)

but this does not guarantee participants' attention. However, we did show our measures had high internal reliability. We included piloted instructions and paused the robot at the junction to ensure participants focused on the cues. There is evidence that video experiments produce meaningful results. Lichtenthäler *et al.* used videos from the point-of-view of a walking human to test navigation algorithms for HRSI [3]. Woods *et al.* showed video results were equivalent to in-person results in an HRSI study [20]. The lack of interactivity in the user study is a limitation, especially with the videos' static point of view. An interactive study would allow for data collection *during* the interaction. Future work should explore these cues with in-person experiments and objective measures such as walking trajectory, as in [14], and human reaction time, as in [15].

## VII. Conclusion

In this research we designed two types of nonverbal robot motion legibility cues, Projected Arrows and Flashing Lights, which had not been directly compared in the reviewed literature. We studied, for the first time, whether these cues should communicate path or goal information, or both sequentially. We also explored turning and straight-ahead Scenarios. Results showed that Projected Arrows should communicate path information and Flashing Lights should communicate goal information. Projected Arrows were communicative in both Robot Movement Scenarios, whereas Flashing Lights were only communicative while turning. Our results also showed that Projected Arrows were more socially acceptable and more comprehensible than Flashing Lights. Based on our findings we recommend the use of Projected Arrows that communicate information about the robot's path. Furthermore, our results showed that the absence of cues was not socially acceptable. Incorporating these motion legibility cues into a mobile robot's design can measurably increase the social acceptance of human-robot spatial interactions in public spaces.